\documentclass[letterpaper]{article} 
\usepackage{aaai23}  
\usepackage{times}  
\usepackage{helvet}  
\usepackage{courier}  
\usepackage[hyphens]{url}  
\usepackage{graphicx} 
\usepackage{natbib}  
\usepackage{caption} 
\usepackage{algorithm}
\usepackage{algorithmic}

\usepackage[utf8]{inputenc} 
\usepackage{booktabs}       
\usepackage{amsfonts}       
\usepackage{nicefrac}       
\usepackage{microtype}      
\usepackage{xcolor}         
\usepackage{xspace}
\usepackage{amsthm}
\usepackage{amsmath,amssymb}
\usepackage{dsfont}
\usepackage{mathtools}

\usepackage{multirow}
\usepackage{makecell}

\usepackage{bm}

\theoremstyle{plain}

\newtheorem{proposition}{Proposition}

\theoremstyle{definition}

\newcommand{\eg}{\emph{e.g.,}\xspace}
\newcommand{\ie}{\emph{i.e.,}\xspace}
\newcommand{\etc}{\emph{etc.}\xspace}

\newcommand{\baby}{P\textsc{lsp}\xspace}

\usepackage{newfloat}
\usepackage{listings}
\DeclareCaptionStyle{ruled}{labelfont=normalfont,labelsep=colon,strut=off} 
\floatstyle{ruled}
\newfloat{listing}{tb}{lst}{}
\floatname{listing}{Listing}
\title{Learning with Partial Labels from Semi-supervised Perspective}
\author {
    Ximing Li\textsuperscript{\rm 1,\rm 2},
    Yuanzhi Jiang\textsuperscript{\rm 1,\rm 2},
    Changchun Li\textsuperscript{\rm 1,\rm 2,}\thanks{Corresponding author.},
    Yiyuan Wang\textsuperscript{\rm 3,\rm 4},
    Jihong Ouyang\textsuperscript{\rm 1,\rm 2}
}
\affiliations {
    \textsuperscript{\rm 1}College of Computer Science and Technology, Jilin University, China\\
    \textsuperscript{\rm 2}Key Laboratory of Symbolic Computation and Knowledge Engineering of MOE, Jilin University, China\\
    \textsuperscript{\rm 3}College of Information Science and Technology, Northeast Normal University, China\\
    \textsuperscript{\rm 4}Key Laboratory of Applied Statistics of MOE, Northeast Normal University, China\\
    liximing86@gmail.com, yzjiang20@mails.jlu.edu.cn, changchunli93@gmail.com, wangyy912@nenu.edu.cn, ouyj@jlu.edu.cn
}

\begin{document}

\maketitle

\begin{abstract}
Partial Label (PL) learning refers to the task of learning from the partially labeled data, where each training instance is ambiguously equipped with a set of candidate labels but only one is valid. Advances in the recent deep PL learning literature have shown that the deep learning paradigms, \eg self-training, contrastive learning, or class activate values, can achieve promising performance. Inspired by the impressive success of deep Semi-Supervised (SS) learning, we transform the PL learning problem into the SS learning problem, and propose a novel PL learning method, namely Partial Label learning with Semi-supervised Perspective (\baby). Specifically, we first form the pseudo-labeled dataset by selecting a small number of reliable pseudo-labeled instances with high-confidence prediction scores and treating the remaining instances as pseudo-unlabeled ones. Then we design a SS learning objective, consisting of a supervised loss for pseudo-labeled instances and a semantic consistency regularization for pseudo-unlabeled instances. We further introduce a complementary regularization for those non-candidate labels to constrain the model predictions on them to be as small as possible. Empirical results demonstrate that \baby significantly outperforms the existing PL baseline methods, especially on high ambiguity levels. Code available: \url{https://github.com/changchunli/PLSP}.

\end{abstract}

\section{Introduction}
\label{1}

\begin{figure}[t]
\centering
\includegraphics[width=0.35\textwidth]{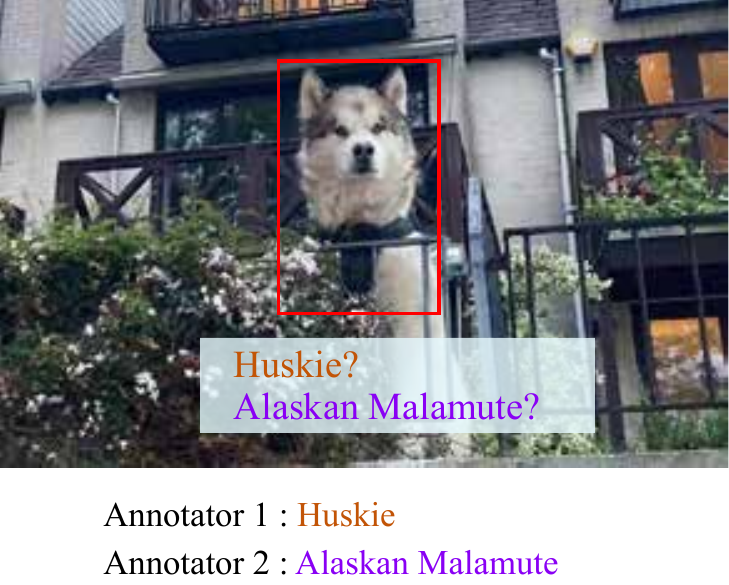}
\caption{An example of PL instances. An \textit{Alaskan Malamute} is in this image, but annotators may also tag it with \textit{Huskie}.}
\label{example}
\end{figure}

During the past decades, modern deep neural networks have gained great success in various domains such as computer vision and natural language processing. Commonly, they are built on the paradigm of supervised learning, which often requires massive training instances with precise labels. However, in many real-world scenarios, the high-quality training instances are intractable to collect, because instance annotation by human-beings is costly and even subject to label ambiguity and noise, potentially resulting in many training data with various noisy supervision \cite{DivideMix2020,P3Mix2022}. Among them, one prevalent noisy challenge is from the partially labeled data, where each training instance is equipped with a set of candidate labels but only one is valid \cite{CLPL2011}. As illustrated in Fig.\ref{example}, for a human annotator it could be difficult to correctly distinguish \textit{Alaskan Malamute} and \textit{Huskie}, so she/he may tend to retain both of them as candidate labels. Due to the popularity of such noisy data in applications, \eg web mining \cite{LCLS2010}, multimedia context analysis \cite{Soccer2013}, and image classification \cite{LALFI2018}, the paradigm of learning from partial labels, formally dubbed as \textbf{P}artial \textbf{L}abel (\textbf{PL}) learning, has recently attracted more attention from the machine learning community \cite{SURE2019,RCCC2020,PRODEN2020,PANGOLIN2020,A2L2CID2021,PiCO2022,PLCR2022}.

Naturally, the main challenge of PL learning lies in the ambiguity of partial labels, because the ground-truth label is unknown and can not be directly accessible to the learning method. Accordingly, the mainstream of PL learning methods concentrates on recovering precise supervised signals from the ambiguous candidate labels. Some two-stage methods refine the candidate labels by label propagation among instance nearest neighbors \cite{IPAL2015,PLLEAF2016,PLLE2019}; and most PL learning methods jointly train the classifier with the refined labels and refine the candidate labels with the classifier predictions \cite{PALOC2018,PLAGGD2019,SDIM2019,PANGOLIN2020,CLAD2021}. Besides them, some deep PL learning methods employ discriminators to recover precise supervision from the candidate labels under the frameworks of GAN \cite{PLGAN2020} and Triple-GAN \cite{A2L2CID2021}. Despite promising performance, the refined labels of these PL learning methods can be still ambiguous and inaccurate for most training instances, resulting in potential performance degradation.

In parallel with PL learning, \textbf{S}emi-\textbf{S}upervised (\textbf{SS}) learning has recently achieved great progress with strong deep neural backbones \cite{Mixmatch2019,ReMixMatch2020,UDA2020,FixMatch2020,FlexMatch2021,S2TCBDD2021}. The recent deep SS learning methods are mainly based on the consistency regularization with the assumption that the classifier tends to give consistent predictions on augmented variants of the same instance. Inspired by them, we revisit the problem that the refined labels of PL learning methods may still be ambiguous for most training instances, and throw the following question: ``\textbf{Whether we can efficiently select a small number of reliable instances with high-confidence pseudo-labels, and then resolve the PL learning task with the strong SS learning techniques?}''


Motivated by this question, we develop a novel PL learning method, namely \textbf{P}artial \textbf{L}abel learning with \textbf{S}emi-supervised \textbf{P}erspective (\textbf{\baby}). Our \baby consists of two stages. In the first stage, we efficiently per-train the classifier by treating all candidate labels equally important, and then select a small number of “reliable” pseudo-labeled instances with high-confidence predictions and treat the remaining instances as pseudo-unlabeled instances. In the second stage, we formulate a SS learning objective to induce the classifier over the pseudo-training dataset. To be specific, we incorporate a consistency regularization with respect to the weakly- and strongly-augmented instances, and draw semantic-transformed features for them to further achieve consistency at the semantic level. To efficiently optimize the objective with semantic-transformed features, we derive an approximation of its expectation form. We conduct extensive experiments to evaluate the effectiveness of \baby on benchmark datasets. Empirical results demonstrate that \baby is superior to the existing PL learning baseline methods, especially on high ambiguity levels.


    
    

\section{Related Work}

\subsection{Partial Label Learning}

There are many PL learning studies based on the shallow frameworks. The early disambiguation-free methods efficiently induce the classifiers by treating all candidate labels equally important \cite{Lost2009,CLPL2011,PLECOC2017}, while in \baby we have also employed this spirit to per-train the classifier to initialize the pseudo-training dataset. Beyond them, the disambiguation methods aim to induce stronger classifiers by refining precise supervision from candidate labels \cite{PALOC2018,LALO2018}. Some two-stage methods first refine the candidate labels by label propagation among instance nearest neighbors \cite{IPAL2015,PLLEAF2016,PLLE2019}. But most disambiguation methods jointly train the classifier with the refined labels and refine the candidate labels with the classifier predictions \cite{SDIM2019,PANGOLIN2020}. However, the refined labels may be also noisy for most training instances.


Inspired by the effectiveness of deep learning and efficiency of stochastic optimization, a number of deep PL learning methods have been recently developed  \cite{PLGAN2020,PRODEN2020,RCCC2020,LW2021,MGPLL2020,A2L2CID2021,IDPLL2021,CAVL2022,PiCO2022}. 
Some deep PL learning methods refine the candidate labels with adversarial training, such as the attempts based on GAN \cite{PLGAN2020} and Triple-GAN \cite{A2L2CID2021}. Most other methods design proper objectives for PL learning. For example, PRODEN \cite{PRODEN2020} optimizes a classifier-consistent objective derived by the assumption that the ground-truth label would achieve the minimal loss among candidate labels; \citet{RCCC2020} propose risk-consistent and classifier-consistent methods with the assumption that the candidate labels are drawn from a uniform distribution; and \citet{LW2021} propose a risk-consistent leveraged weighted loss with label-specific candidate label sampling. However, those methods highly rely on their prior assumptions. Besides, the recent deep PL learning method PiCO \cite{PiCO2022} borrows the idea of contrastive learning to keep the consistence between the augmented versions of each instance. In contrast to PiCO, we also employ a consistency regularization with respect to augmented instances but we further draw semantic-transformed features for them to achieve a new semantic consistency regularization.


\subsection{Semi-Supervised Learning}

The recent deep SS learning methods are mainly based on the consistency regularization \cite{2016temporal,MeanT2017,VAT2019,Mixmatch2019,ReMixMatch2020,UDA2020,FixMatch2020,FlexMatch2021}.
It is built on a simple concept that the classifier should give consistent predictions on an unlabeled instance and its perturbed version. To conduct this idea, many perturbation methods are adopted, such as the virtual adversarial training used in VAT \cite{VAT2019} and the mixup technique adopted by MixMatch \cite{Mixmatch2019}. With the data augmentation technique popular, the previous arts \cite{ReMixMatch2020,UDA2020,FixMatch2020,FlexMatch2021} keep the classifier predictions on the weakly- and strongly-augmented variants of an unlabeled instance to be consistent, and empirically achieve impressive performance. The main data augmentation techniques used in these methods are at the pixel-level, such as flip-and-shift, Cutout \cite{Cutout2017}, AutoAugment \cite{AutoAugment2019}, CTAugment \cite{ReMixMatch2020}, and RandAugment \cite{RandAugment2020} \etc Complementary to these pixel-level augmentations, \citet{ISDA2022} design a semantic-level data augmentation method motivated by the linear characteristic of deep features. In \baby, we employ both pixel-level and semantic-level data augmentations, and design a semantic consistency regularization for PL learning by performing semantic-level transformations on both weakly- and strongly-augmented variants of an instance.

\section{The Proposed \baby Approach}
\label{3}

In this section, we introduce the proposed PL learning method, namely \textbf{P}artial \textbf{L}abel learning with \textbf{S}emi-supervised \textbf{P}erspective (\textbf{\baby}).

\paragraph{Problem formulation of PL learning}
We now formulate the problem of PL learning. Let $\mathcal{X} \subseteq \mathbb{R}^d$ be a $d$-dimensional feature space and $\mathcal{Y} = \{1,\cdots,l\}, l \geq 2$ be the label space. In the context of PL learning, we are given by a training dataset consisting of $n$ instances, denoted by $\Omega=\{(\mathbf{x}_i,C_i)\}_{i=1}^{n}$. For each instance, $\mathbf{x}_i\in\mathcal{X}$ and $C_i \subseteq \mathcal{C}$ are its feature vector and corresponding candidate label set, respectively, where $\mathcal{C} = \{\mathcal{P}(\mathcal{Y})\setminus\emptyset\setminus\mathcal{Y}\}$ is the power set of $\mathcal{Y}$ except for the empty set and the whole label set. Specially, the single ground-truth label of each instance is unknown and must be concealed in its corresponding candidate label set. The goal of PL learning is to train a classifier $f(\cdot \:;\mathbf{\Theta})$, parameterized by $\mathbf{\Theta}$, from such noisy training dataset $\Omega$.

\subsection{Overview of \baby}
The main idea of \baby is to transform the PL learning problem into the SS learning problem, and then induce the classifier by leveraging the well-established SS learning paradigms. Specifically, we first select a small number of ``reliable'' partial-labeled instances (\ie $m\ll n$) from $\Omega$ according to their predicted scores, \eg class activation values \cite{CAVL2022}, and form a pseudo-labeled instance set $\Omega_l = \{(\mathbf{x}_{p(i)},y_{p(i)}) \in \mathcal{X} \times \mathcal{C}_{p(i)}\}_{i=1}^m$, where the subscript $p(i)$ denotes the mapping function of instance index and $y_{p(i)}$ is the corresponding high-confidence pseudo label. We treat the remaining instances as pseudo-unlabeled instances, denoted by $\Omega_u$. Accordingly, we can further treat the pseudo-training dataset $\{\Omega_l,\Omega_u\}$ as a training dataset of SS learning, so as to train a classifier from it by leveraging the following well-established objective of SS learning:
\begin{equation}
\label{eq-3-1}
    \mathcal{L}(\{\Omega_l,\Omega_u\};\mathbf{\Theta}) = \mathcal{L}_l(\Omega_l;\mathbf{\Theta}) + \mathcal{R}_u(\Omega_u;\mathbf{\Theta}),
\end{equation}
where $\mathcal{L}_l$ is the pseudo-supervised loss with respect to $\Omega_l$; and $\mathcal{R}_u$ the regularization with respect to $\Omega_u$ (\eg consistency regularization). In the following, we will introduce the details of forming the pseudo-training dataset $\{\Omega_l,\Omega_u\}$ and constructing SS learning loss over $\{\Omega_l,\Omega_u\}$ for PL learning, and then show the specific objective of \baby as well as the full training process. 

\subsection{Forming the Pseudo-Training Dataset $\{\Omega_l,\Omega_u\}$}

To form the pseudo-labeled instance set $\Omega_l$, we pre-train the classifier $f(\cdot\:;\mathbf{\Theta})$ by using a simple disambiguation-free PL learning loss, where all candidate labels are treated equally:
\begin{equation}
    \label{eq-3-0}
    \mathcal{L}_{df}(\Omega;\mathbf{\Theta})=\frac{1}{\lvert \Omega \rvert}\sum_{(\mathbf{x}_i,C_i)\in\Omega}\frac{1}{\lvert C_i \rvert}\sum_{j\in C_i} -\log p_{ij}
\end{equation}
where $\mathbf{p}_i=[p_{ij}]_{j\in\mathcal{Y}}^{\top}$ is the classifier prediction of instance $\mathbf{x}_i$, and
$p_{ij}=e^{z_{ij}}/\sum_{j'\in\mathcal{Y}} e^{z_{ij'}},\;\mathbf{z}_i=f(\mathbf{x}_i;\mathbf{\Theta})$.
With the per-trained classifier $f(\cdot\:;\mathbf{\widetilde \Theta})$,\footnote{This pre-training process can be very efficient and converge within a few epochs.} we select a small number of ``reliable'' pseudo-labeled instances to form $\Omega_l$. Specifically, for each instance $(\mathbf{x}_i,C_i)\in\Omega$, we assign the candidate label with the highest class activation value (CAV) \cite{CAVL2022} as its pseudo label:
\begin{equation*}
y_i=\mathop{\arg\max}\limits_{j\in C_i} v_{ij},\;\; v_{ij}=\widetilde z_{ij}\lvert \widetilde z_{ij}-1\rvert,\;\;\mathbf{\widetilde z}_i=f(\mathbf{x}_i;\mathbf{\widetilde \Theta}).
\end{equation*}
For each class $j\in\mathcal{Y}$, we construct its pseudo-labeled instance set $\Omega_l^j$ by choosing instances with the top-$k$ CAVs of class $j$:
\begin{equation*}
\Omega_l^j=\bigl\{(\mathbf{x}_i,y_i)|i \in \text{TopK}(\{v_{ij}|(\mathbf{x}_i,C_i)\in\Omega,y_i=j\})\bigr\},
\end{equation*}
where, as its name suggests, $\text{TopK}(\cdot)$ outputs the index set of instances with the top-$k$ CAVs. Accordingly, the pseudo-labeled set $\Omega_l$ can be formed as follows:\footnote{The total number of pseudo-labeled instances $m= k \times l$.}
\begin{equation}
\label{eq-3-4}
\Omega_l=\bigcup_{j\in\mathcal{Y}}\Omega_l^j,
\end{equation}
and the remaining instances can constitute the pseudo-unlabeled instance set $\Omega_u$ as follows:
\begin{equation}
\label{eq-3-4-1}
\Omega_u=\{(\mathbf{x}_i,C_i)|(\mathbf{x}_i,C_i)\in\Omega,(\mathbf{x}_i,y_i)\notin\Omega_l\}.
\end{equation}

\subsection{Forming the SS Learning Loss over $\{\Omega_l,\Omega_u\}$}
Given $\{\Omega_l,\Omega_u\}$, we continue to optimize the per-trained classifier $f(\cdot\:;\mathbf{\widetilde \Theta})$ by using the SS learning loss of \baby, including the pseudo-supervised loss $\mathcal{L}_l$ for $\Omega_l$ and the regularization term $\mathcal{R}_u$ for $\Omega_u$. 

\paragraph{Pseudo-supervised loss.} 
We can treat $\Omega_l$ as a labeled dataset, and directly formulate the specific pseudo-supervised loss as follows:
\begin{equation}
\label{eq-3-5}
\mathcal{L}_l(\Omega_l;\mathbf{\Theta})=\frac{1}{\lvert\Omega_l\rvert}\sum_{(\mathbf{x}_i,y_i)\in\Omega_l}-\log p_{iy_i}.
\end{equation}

\paragraph{Regularizing the pseudo-unlabeled instances}
Inspired by the impressive success of the consistency regularization in SS learning \cite{UDA2020,FixMatch2020,FlexMatch2021}, we employ it to regularize the pseudo-unlabeled instances. Specifically, for each instance within $\Omega_u$, we first generate its weakly- and strongly-augmented variants with the wide-used pixel-level data augmentation tricks,\footnote{These data augmentations tricks include flip-and-shift, Cutout \cite{Cutout2017}, AutoAugment \cite{AutoAugment2019}, CTAugment \cite{ReMixMatch2020}, and RandAugment \cite{RandAugment2020} \etc We will introduce their implementation details in the experiment part.} and then constrain their corresponding prediction scores to be consistent. Formally, for each pseudo-unlabeled instance $(\mathbf{x}_i, C_i)\in\Omega_u$, let its weakly- and strongly-augmented variants denote by $\mathbf{x}_i^w=\alpha(\mathbf{x}_i)$ and $\mathbf{x}_i^s=\mathcal{A}(\mathbf{x}_i)$, respectively. Its corresponding consistency regularization term can be written as follows:
\begin{equation}
\label{eq-3-7}
\mathfrak{R}_u((\mathbf{x}_i,C_i);\mathbf{\Theta})= h(\mathbf{\widehat p}_i^w)\text{KL}(\mathbf{\widehat p}_i||\mathbf{p}_i^s),
\end{equation}
where $\text{KL}(\cdot||\cdot)$ denotes the KL-divergence. More specially, $\mathbf{\widehat p}_i=[\widehat p_{ij}]_{j\in\mathcal{Y}}^{\top}$ is the pseudo-target approximated on the weakly-augmented variant $\mathbf{x}_i^w$:
\begin{align*}
    \widehat{p}_{ij} & =\frac{\mathds{1}(j\in C_i)\widehat{p}_{ij}^w}{\sum_{j'\in \mathcal{Y}}\mathds{1}(j'\in C_i)\widehat{p}_{ij'}^w},\\
    \widehat{p}_{ij}^w & =\frac{e^{\widehat{z}_{ij}^w}}{\sum_{j'\in\mathcal{Y}} e^{\widehat{z}_{ij'}^w}},  \quad \:\:
    \mathbf{\widehat z}_i^w=f(\mathbf{x}_i^w;\mathbf{\widehat \Theta}),
\end{align*}
where $\mathbf{\widehat \Theta}$ is the fixed copy of the current parameters $\mathbf{\Theta}$;
$\mathbf{p}_i^s = [p^s_{ij}]_{j\in\mathcal{Y}}^{\top}$ is the classifier prediction on the strongly-augmented variant $\mathbf{x}_i^s$, and $p_{ij}^s =e^{z_{ij}^s}/\sum_{j'\in\mathcal{Y}} e^{z_{ij'}^s},\;\mathbf{z}_i^s=f(\mathbf{x}_i^s;\mathbf{\Theta})$;
$h(\mathbf{\widehat p}_i^w)$ is an indicator function used to retain high-confidence pseudo-unlabeled instances in this consistency regularization term, specifically defined as follows: 
\begin{equation*}
    h(\mathbf{\widehat p}_i^w) = \mathds{1}\left(\Bigl(\mathop{\max}\limits_{j\in\mathcal{Y}} \widehat{p}_{ij}^w\geq\tau \Bigr) \wedge \Bigl(\mathop{\arg\max}\limits_{j\in\mathcal{Y}} \widehat{p}_{ij}^w\in C_i\Bigr)\right),
\end{equation*}
where $\tau\in(0.5,1.0]$ is the confidence threshold.
Accordingly, the overall consistency regularization over $\Omega_u$ is stated as:
\begin{equation}
\label{eq-3-6}
\mathcal{R}_{u}(\Omega_u;\mathbf{\Theta})=\frac{1}{\lvert\Omega_u\rvert}\sum_{(\mathbf{x}_i,C_i)\in\Omega_u}\mathfrak{R}_u((\mathbf{x}_i,C_i);\mathbf{\Theta}).
\end{equation}
Besides, inspired by that the deep feature space usually is linear and includes some meaningful semantic directions \cite{ISDA2019,ISDA2022}, we construct semantic-level transformations based on those semantic directions, which is complementary to the pixel-level transformations, and perform semantic consistency regularization on them, so as to further regularize the classifier $f(\cdot\:;\mathbf{\Theta})$ in the semantic level. Let the classifier $f(\cdot\:;\mathbf{\Theta})=\mathbf{W}^{\top}g(\cdot\:;\mathbf{\Phi})$, $g(\cdot\:;\mathbf{\Phi})$ be the deep feature extractor, and $\mathbf{W}=[\mathbf{w}_j]_{j\in\mathcal{Y}}^{\top}$ be parameters of the last full-connected predictive layer. Specifically, we suppose that those semantic directions are drawn from a set of label-specific zero-mean Gaussian distributions $\{\mathcal{N}(\mathbf{0},\lambda\mathbf{\Sigma}_j)\}_{j\in\mathcal{Y}}$, where $\lambda>0$ controls the strength of semantic transformations. Given the known labels, we can apply those sampled label-specific semantic directions on the deep features of instances to construct semantic-level transformations. Thanks to the properties of Gaussian distribution, we can draw the semantic-transformed feature of any instance $(\mathbf{x}_i, y_i)$ as:
\begin{equation}
\label{eq-3-14}
    \underline{\mathbf{a}}_i\sim\mathcal{N}(\mathbf{a}_i,\lambda\mathbf{\Sigma}_{y_i}),\;\mathbf{a}_i=g(\mathbf{x}_i;\mathbf{\Phi}).
\end{equation}
Nevertheless, the true labels of pseudo-unlabeled instances within $\Omega_u$ are totally unknown. For each instance $(\mathbf{x}_i, C_i)\in\Omega_u$,  we approximate its pseudo label $\widehat{y}_i$ with the CAVs on its weakly-augmented variant $\mathbf{x}_i^w$ as:
\begin{align*}
    &\widehat{y}_i=\mathop{\arg\max}\limits_{j\in\mathcal{C}_i}\widehat{v}_{ij},\;\widehat{v}_{ij}=\widehat{z}_{ij}\lvert \widehat{z}_{ij}-1\rvert,\;\mathbf{\widehat z}_i=f(\mathbf{x}_i^w;\mathbf{\widehat\Theta}).
\end{align*}
We can then draw the semantic-transformed features of its weakly- and strongly-augmented variants by applying Eq.\eqref{eq-3-14}.
Drawing $K$ semantic-transformed features for each augmented variant, the consistency regularization term in Eq.\eqref{eq-3-7} can be rewritten as the following semantic consistency regularization term:
\begin{align}
\label{eq-3-8}
&\mathfrak{R}_u^K((\mathbf{x}_i,C_i);\mathbf{\Theta})=
\frac{1}{K^2}\sum_{k_1,k_2=1}^K h(\mathbf{\widehat p}_i^{w,k_1})\text{KL}(\mathbf{\widehat p}_i^{k_1}||\mathbf{p}_i^{s,k_2}), \nonumber \\
&\quad\textbf{s.t.}\quad\underline{\mathbf{\widehat a}}_i^{w,k_1}\sim\mathcal{N}(\mathbf{\widehat a}_i^w,\lambda\mathbf{\Sigma}_{\widehat{y}_i}),\;\mathbf{\widehat a}_i^w=g(\mathbf{x}_i^w;\mathbf{\widehat \Phi}), \nonumber \\
&\quad\quad\quad\:\underline{\mathbf{a}}_i^{s,k_2}\sim\mathcal{N}(\mathbf{a}_i^s,\lambda\mathbf{\Sigma}_{\widehat{y}_i}),\;\mathbf{a}_i^s=g(\mathbf{x}_i^s;\mathbf{\Phi}),
\end{align}
where
\begin{equation*}
    \widehat{p}_{ij}^{w,k_1}=\frac{e^{\widehat{z}_{ij}^{w,k_1}}}{\sum_{j'\in\mathcal{Y}} e^{\widehat{z}_{ij'}^{w,k_1}}},\quad \mathbf{\widehat z}_i^{w,k_1}=\mathbf{\widehat W}^{\top}\underline{\mathbf{\widehat a}}_i^{w,k_1};
\end{equation*}
\begin{equation*}
    p_{ij}^{s,k_2}=\frac{e^{z_{ij}^{s,k_2}}}{\sum_{j'\in\mathcal{Y}} e^{z_{ij'}^{s,k_2}}},\quad \mathbf{z}_i^{s,k_2}=\mathbf{W}^{\top}\underline{\mathbf{a}}_i^{s,k_2},
\end{equation*}
and further $\mathbf{\widehat p}_i^{k_1} = [\widehat p^{k_1}_{ij}]_{j\in\mathcal{Y}}^{\top}$ is calculated as follows:
\begin{equation*}
    \widehat{p}_{ij}^{k_1}=\frac{\mathds{1}(j\in C_i)\widehat{p}_{ij}^{w,k_1}}{\sum_{j'\in \mathcal{Y}}\mathds{1}(j'\in C_i)\widehat{p}_{ij'}^{w,k_1}}.
\end{equation*}
To avoid inefficiently sampling, we consider the expectation of Eq.\eqref{eq-3-8} with all possible semantic-transformed features:
\begin{equation}
\label{eq-3-9}
    \mathfrak{R}_u^{\infty}((\mathbf{x}_i,C_i);\mathbf{\Theta})=\mathbb{E}_{\underline{\mathbf{\widehat a}}_i^{w,k_1},\underline{\mathbf{ a}}_i^{s,k_2}}[h(\mathbf{\widehat p}_i^{w,k_1})\text{KL}(\mathbf{\widehat p}_i^{k_1}||\mathbf{p}_i^{s,k_2})].
\end{equation}
Unfortunately, it is intractable to optimize Eq.\eqref{eq-3-9} in its exact form. Alternatively, we derive an easy-to-compute upper bound $\overline{\mathfrak{R}}_u^{\infty}((\mathbf{x}_i,C_i);\mathbf{\Theta})$ given in the following proposition. Finally, the consistency regularization over $\Omega_u$ in Eq.\eqref{eq-3-6} is rewritten below:
\begin{equation}
    \label{eq-3-11}
    \overline{\mathcal{R}}_{u}(\Omega_u;\mathbf{\Theta})=\frac{1}{\lvert\Omega_u\rvert}\sum_{(\mathbf{x}_i,C_i)\in\Omega_u}\overline{\mathfrak{R}}_u^{\infty}((\mathbf{x}_i,C_i);\mathbf{\Theta}).
\end{equation}
\begin{proposition}
Suppose that $\underline{\mathbf{\widehat a}}_i^{w,k_1}\sim\mathcal{N}(\mathbf{\widehat a}_i^w,\lambda\mathbf{\Sigma}_{\widehat{y}_i})$ and $\underline{\mathbf{ a}}_i^{s,k_2}\sim\mathcal{N}(\mathbf{a}_i^s,\lambda\mathbf{\Sigma}_{\widehat{y}_i})$. Then we have an upper bound for $\mathfrak{R}_u^{\infty}((\mathbf{x}_i,C_i);\mathbf{\Theta})$ given by
\begin{align}
\label{eq-3-10}
&\mathfrak{R}_u^{\infty}((\mathbf{x}_i,C_i);\mathbf{\Theta})\leq h(\mathbf{\underline{\widehat p}}_i^w)\textup{KL}(\mathbf{\underline{\widehat p}}_i||\mathbf{\underline{p}}_i^s) \triangleq \overline{\mathfrak{R}}_u^{\infty}((\mathbf{x}_i,C_i);\mathbf{\Theta}) \nonumber \\
&\quad\textbf{\textup{s.t.}}\;\;\mathbf{\widehat a}_i^w=g(\mathbf{x}_i^w;\mathbf{\widehat \Phi}),\;
\mathbf{a}_i^s=g(\mathbf{x}_i^s;\mathbf{\Phi}),
\end{align}
where $\underline{\widehat{p}}_{ij}^w=\frac{1}{-l+\sum_{j'\in\mathcal{Y}}1/\Phi\Bigl(\frac{\beta\mathbf{\widehat u}_{jj'}^{\top}\mathbf{\widehat a}_i^w}{(1+\lambda\beta^2\mathbf{\widehat u}_{jj'}^{\top}\mathbf{\Sigma}_{\widehat{y}_i}\mathbf{\widehat u}_{jj'})^{1/2}}\Bigr)}$, 
$\underline{\widehat p}_{ij}= \frac{\mathds{1}(j\in C_i)\underline{\widehat{p}}_{ij}^w}{\sum_{j'\in \mathcal{Y}}\mathds{1}(j'\in C_i)\underline{\widehat{p}}_{ij'}^w}$, $\underline{p}_{ij}^s=\frac{e^{\mathbf{w}_j^{\top}\mathbf{a}_i^s}}{\sum_{j'\in\mathcal{Y}}e^{\mathbf{w}_{j'}^{\top}\mathbf{a}_i^s+\frac{\lambda}{2}\mathbf{u}_{j'j}^{\top}\mathbf{\Sigma}_{\widehat{y}_i}\mathbf{u}_{j'j}}}$,
$\mathbf{\widehat u}_{jj'}=\mathbf{\widehat w}_j-\mathbf{\widehat w}_{j'}$, $\mathbf{u}_{j'j}=\mathbf{w}_{j'}-\mathbf{w}_j$, and $\Phi(z)=\frac{1}{\sqrt{2\pi}}\int_{-\infty}^ze^{-t^2/2}dt$ is the cumulative distribution function of the standard normal distribution $\mathcal{N}(0,1)$.
\end{proposition}

\subsection{Objective of \baby and Iterative Training Summary}
We summarize the overall objective of \baby, and clarify the training details in following.

\paragraph{Objective of \baby.} 
Besides the aforementioned pseudo-supervised loss $\mathcal{L}_l$ and regularization $\mathcal{R}_u$, we also incorporate a complementary loss over $\Omega_u$ to minimize the predictions of non-candidate labels:
\begin{equation}
\label{eq-3-12}
    \mathcal{L}_{cl}(\Omega_u;\mathbf{\Theta})=\frac{1}{\lvert\Omega_u\rvert}\sum_{(\mathbf{x}_i,C_i)\in\Omega_u}\sum_{j\notin C_i}-log(1-p_{ij}).
\end{equation}
And we also improve $\mathcal{L}_l$ and $\mathcal{L}_{cl}$ with the semantic-level transformation Eq.\eqref{eq-3-14}, then obtain their corresponding upper bounds according to Proposition 1, given by:
\begin{align}
    &\overline{\mathcal{L}}_l(\Omega_l;\mathbf{\Theta})=\frac{1}{\lvert \Omega_l \rvert}\sum_{(\mathbf{x}_i,y_i)\in\Omega_l}-\log\underline{p}_{iy_i}, \label{eq-3-15} \\
    &\overline{\mathcal{L}}_{cl}(\Omega_u;\mathbf{\Theta})=\frac{1}{\lvert \Omega_u \rvert}\sum_{(\mathbf{x}_i,C_i)\in\Omega_u}\sum_{j\notin C_i}-\log(1-\underline{p}_{ij}), \label{eq-3-16}
\end{align}
where
\begin{align*}
    &\underline{p}_{iy_i}=\frac{e^{\mathbf{w}_{y_i}^{\top}\mathbf{a}_i}}{\sum_{j'\in\mathcal{Y}}e^{\mathbf{w}_{j'}^{\top}\mathbf{a}_i+\frac{\lambda}{2}\mathbf{u}_{j'y_i}^{\top}\mathbf{\Sigma}_{y_i}\mathbf{u}_{j'y_i}}},\;\mathbf{a}_i=g(\mathbf{x}_i,\mathbf{\Phi}), \\
    &\underline{p}_{ij}=\frac{e^{\mathbf{w}_j^{\top}\mathbf{a}_i}}{\sum_{j'\in\mathcal{Y}}e^{\mathbf{w}_{j'}^{\top}\mathbf{a}_i+\frac{\lambda}{2}\mathbf{u}_{j'j}^{\top}\mathbf{\Sigma}_{\widehat{y}_i}\mathbf{u}_{j'j}}},\;\mathbf{a}_i=g(\mathbf{x}_i,\mathbf{\Phi}),
\end{align*}
and $\mathbf{u}_{j'y_i}=\mathbf{w}_{j'}-\mathbf{w}_{y_i}$. Accordingly, the final objective of \baby can be reformulated as:
\begin{align}
    \label{eq-3-17}
    &\mathcal{L}(\{\Omega_l,\Omega_u\};\mathbf{\Theta})= \nonumber \\
    &\quad\quad\gamma\left(\overline{\mathcal{L}}_l(\Omega_l;\mathbf{\Theta})+\overline{\mathcal{R}}_u(\Omega_u;\mathbf{\Theta})\right)+\overline{\mathcal{L}}_{cl}(\Omega_u;\mathbf{\Theta}),
\end{align}
where $\gamma>0$ is the hyper-parameter to balance the SS learning loss and complementary loss.

\paragraph{Update of label-specific covariance matrices $\{\mathbf{\Sigma}_j\}_{j\in\mathcal{Y}}$.} Following \cite{ISDA2019,ISDA2022}, we approximate $\{\mathbf{\Sigma}_j\}_{j\in\mathcal{Y}}$ with pseudo-labeled instances by counting statistics from all mini-batches incrementally. For each $\mathbf{\Sigma}_j$ in the $c$-th iteration, it can be updated as follows:
\begin{align}
\label{eq-3-18}
    &\!\!\!\!\mathbf{\Sigma}_j^{(c)}=\frac{m_j^{(c-1)}\mathbf{\Sigma}_j^{(c-1)}+{m'}_j^{(c)}{\mathbf{\Sigma}'}_j^{(c)}}{m_j^{(c-1)}+{m'}_j^{(c)}} \nonumber \\
    &\!\!\!+\frac{m_j^{(c-1)}{m'}_j^{(c)}(\bm{\mu}_j^{(c-1)}-{\bm{\mu}'}_j^{(c)})(\bm{\mu}_j^{(c-1)}-{\bm{\mu}'}_j^{(c)})^{\top}}{(m_j^{(c-1)}+{m'}_j^{(c)})^2},
\end{align}
\begin{align*}
    \bm{\mu}_j^{(c)}=\frac{m_j^{(c-1)}\bm{\mu}_j^{(c-1)}+{m'}_j^{(c)}{\bm{\mu}'}_j^{(c)}}{m_j^{(c-1)}+{m'}_j^{(c)}},\;
    m_j^{(c)}=m_j^{(c-1)}+{m'}_j^{(c)},
\end{align*}
where ${\bm{\mu}'}_j^{(c)}$ and ${\mathbf{\Sigma}'}_j^{(c)}$ are the mean and covariance matrix of features within class $j$ in $c$-th mini-batch, respectively; $m_j^{(c)}$ the total number of pseudo-labeled instances belonging to class $j$ in all $c$ mini-batches and ${m'}_j^{(c)}$ the number of pseudo-labeled instances belonging to class $j$ in $c$-th mini-batch.

\paragraph{Adjusting the SS learning loss weight $\gamma$.} In the early training stage, the SS learning loss may be less accurate. To fix issue, We dynamically adjust the SS learning loss weight $\gamma$ by a non-decreasing function $\gamma=\min\{\frac{t}{T}\gamma_0,\gamma_0\}$ with respect to the epoch number $t$, where $\gamma_0$ is the maximum weight, and $T$ the maximum number of SS training epochs.

\paragraph{Adjusting the confidence threshold $\tau$.} We employ the curriculum pseudo labeling \cite{FlexMatch2021} to adjust $\tau$. For each class $j$, its value at $c$-th iteration is calculated by:
\begin{align*}
    &\tau_c(j)=\eta_c(j)\cdot\tau_0,\;\eta_c(j)=\frac{\sigma_c(j)}{\mathop{\max}_{j'\in\mathcal{Y}}\sigma_c(j')}, \\
    &\sigma_c(j)=\sum_{(\mathbf{x}_i,C_i)\in\Omega_u}h(\underline{\mathbf{\widehat p}}_i^w)\cdot\mathds{1}(\widehat{y}_i=j),
\end{align*}
where $\tau_0$ is the maximum confidence threshold.

\paragraph{Adjusting the transformation strength $\lambda$.} Following \cite{ISDA2019,ISDA2022}, we dynamically adjust the transformation strength $\lambda$ with a non-decreasing function $\lambda=\min\{\frac{t}{T}\lambda_0,\lambda_0\}$ with respect to the epoch number $t$, where $\lambda_0$ is the maximum transformation strength, so as to reduce the negative impact of the low-quality estimations of covariance matrices in the early training stage.

\paragraph{Iterative training summary.}
In practice, to prevent the error memorization and reduce the time cost, we update $\{\Omega_l,\Omega_u\}$ with the current predictions per-epoch. The classifier parameters $\mathbf{\Theta}$ are optimized by using the stochastic optimization with SGD. Overall, the iterative training procedure of \baby is summarized in \textit{Algorithm \ref{Alg1}}.

\setlength{\textfloatsep}{5pt}
\begin{algorithm}[t]
  \caption{Training procedure of \baby}
  \label{Alg1}
  \textbf{Input}: $\Omega$: PL training dataset $\Omega=\{(\mathbf{x}_i,C_i)\}_{i=1}^{n}$; $m$: number of pseudo-labeled instances; $\gamma_0$: SS learning loss weight; $\tau_0$: confidence threshold; $\lambda_0$: semantic transformation strength; \\
  \textbf{Output}: $\mathbf{\Theta}$: classifier parameters
  \begin{algorithmic}[1]
    \STATE Initialize the classifier parameters $\mathbf{\Theta}=\{\mathbf{\Phi},\mathbf{W}\}$; 
    \FOR[\% Pre-training stage \%]{$t=0$ \textbf{to} $T_0$}
    \FOR{$c=0$ \textbf{to} $I$}
    \STATE Sample a mini-batch $\{(\mathbf{x}_i,C_i)\}_{i=1}^B$ from $\Omega$;
    \STATE Compute $\mathcal{L}_{df}$ according to Eq.\eqref{eq-3-0};
    \STATE Update $\mathbf{\Theta}$ with SGD;
    \ENDFOR
    \ENDFOR
    \FOR[\% SS training stage \%]{$t=0$ \textbf{to} $T$}
    \STATE Construct pseudo-training dataset $\{\Omega_l,\Omega_u\}$ according to Eqs.\eqref{eq-3-4} and \eqref{eq-3-4-1};
    \FOR{$c=0$ \textbf{to} $I$}
    \STATE Sample a mini-batch $\{(\mathbf{x}_i,y_i)\}_{i=1}^{B_l}$ from $\Omega_l$ and a mini-batch $\{(\mathbf{x}_i^w, \mathbf{x}_i^s, C_i)\}_{i=1}^{B_u}$ from $\Omega_u$ with $\alpha(\cdot)$ and $\mathcal{A}(\cdot)$;
    \STATE Compute $\mathbf{a}_i=g(\mathbf{x}_i;\mathbf{\Phi}), \mathbf{a}_i^w=g(\mathbf{x}_i^w;\mathbf{\Phi}), \mathbf{a}_i^s=g(\mathbf{x}_i^s;\mathbf{\Phi})$;
    \STATE Estimate covariance matrices $\{\mathbf{\Sigma}_j\}_{j\in\mathcal{Y}}$ according to Eq.\eqref{eq-3-18};
    \STATE Compute $\mathcal{L}$ according to Eq.\eqref{eq-3-17};
    \STATE Update $\mathbf{\Theta}$ with SGD;
    \ENDFOR
    \ENDFOR
  \end{algorithmic}
\end{algorithm}

\section{Experiment}
\label{4}


\begin{table*}[ht]
  \centering
  \caption{Empirical results (mean$\pm$std) on Fashion-MNIST (FMNIST) and CIFAR-10 with different data generation strategies and ambiguity levels: (\uppercase\expandafter{\romannumeral1}) USS; (\uppercase\expandafter{\romannumeral2}) FPS ($q$=0.3); (\uppercase\expandafter{\romannumeral3}) FPS ($q$=0.5); (\uppercase\expandafter{\romannumeral4}) FPS ($q$=0.7). The highest scores are indicated in \textbf{bold}. The notation ``$\ddagger$'' indicates that the performance gain of \baby is statistically significant (paired sample t-tests) at $0.01$ level.}
  \label{table1}
  \footnotesize
  \resizebox{0.98\linewidth}{!}{
  \begin{tabular}{p{36pt}<{\centering}|p{44pt}<{\centering}||p{48pt}<{\centering}p{48pt}<{\centering}p{48pt}<{\centering}p{48pt}<{\centering}p{48pt}<{\centering}p{48pt}<{\centering}p{48pt}<{\centering}}
  \Xhline{1.5pt}
  Metric &Dataset &\textbf{CC} &\textbf{RC} &\textbf{PRODEN} &\textbf{LWS} &\textbf{CAVL} &\textbf{PiCO} &\textbf{\baby} \\
  \hline\hline
  \multicolumn{9}{c}{(\uppercase\expandafter{\romannumeral1}) USS} \\
  \hline
  \multirow{2}{*}{Macro-F1} &FMNIST &{0.879$\pm$0.001}$^{\ddagger}$ &{0.893$\pm$0.001} &{0.891$\pm$0.006} &{0.881$\pm$0.007}$^{\ddagger}$ &{0.882$\pm$0.004}$^{\ddagger}$ &\textbf{0.907$\pm$0.001} &{0.897$\pm$0.003} \\
  ~ &CIFAR-10 &{0.745$\pm$0.006}$^{\ddagger}$ &{0.787$\pm$0.003}$^{\ddagger}$ &{0.796$\pm$0.003}$^{\ddagger}$ &{0.781$\pm$0.007}$^{\ddagger}$ &{0.748$\pm$0.051}$^{\ddagger}$ &{0.869$\pm$0.001}$^{\ddagger}$  &\textbf{0.889$\pm$0.003} \\
  \hline
  \multirow{2}{*}{Micro-F1} &FMNIST &{0.880$\pm$0.001}$^{\ddagger}$ &{0.893$\pm$0.001} &{0.892$\pm$0.003} &{0.883$\pm$0.005}$^{\ddagger}$ &{0.883$\pm$0.004}$^{\ddagger}$ &\textbf{0.907$\pm$0.001} &{0.897$\pm$0.002} \\
  ~ &CIFAR-10 &{0.747$\pm$0.005}$^{\ddagger}$ &{0.788$\pm$0.003}$^{\ddagger}$ &{0.796$\pm$0.003}$^{\ddagger}$ &{0.782$\pm$0.004}$^{\ddagger}$ &{0.756$\pm$0.038}$^{\ddagger}$ &{0.870$\pm$0.001}$^{\ddagger}$ &\textbf{0.889$\pm$0.003}  \\
  \hline\hline
  \multicolumn{9}{c}{(\uppercase\expandafter{\romannumeral2}) FPS ($q=0.3$)} \\
  \hline
  \multirow{2}{*}{Macro-F1} &FMNIST &{0.883$\pm$0.003}$^{\ddagger}$ &{0.893$\pm$0.001} &{0.894$\pm$0.003} &{0.888$\pm$0.005} &{0.887$\pm$0.003} &\textbf{0.909$\pm$0.003} &{0.894$\pm$0.002} \\
  ~ &CIFAR-10 &{0.769$\pm$0.005}$^{\ddagger}$ &{0.803$\pm$0.003}$^{\ddagger}$ &{0.801$\pm$0.005}$^{\ddagger}$ &{0.802$\pm$0.005}$^{\ddagger}$ &{0.796$\pm$0.004}$^{\ddagger}$ &{0.880$\pm$0.003}$^{\ddagger}$  &\textbf{0.898$\pm$0.002} \\
  \hline
  \multirow{2}{*}{Micro-F1} &FMNIST &{0.883$\pm$0.003} &{0.894$\pm$0.001} &{0.895$\pm$0.002} &{0.889$\pm$0.004} &{0.887$\pm$0.003} &\textbf{0.909$\pm$0.003} &{0.894$\pm$0.002} \\
  ~ &CIFAR-10 &{0.769$\pm$0.004}$^{\ddagger}$ &{0.803$\pm$0.003}$^{\ddagger}$ &{0.801$\pm$0.005}$^{\ddagger}$ &{0.803$\pm$0.004}$^{\ddagger}$ &{0.796$\pm$0.004}$^{\ddagger}$ &{0.880$\pm$0.003}$^{\ddagger}$ &\textbf{0.898$\pm$0.002} \\
  \hline\hline
  \multicolumn{9}{c}{(\uppercase\expandafter{\romannumeral3}) FPS ($q=0.5$)} \\
  \hline
  \multirow{2}{*}{Macro-F1} &FMNIST &{0.881$\pm$0.001} &{0.890$\pm$0.002} &{0.891$\pm$0.004} &{0.884$\pm$0.006} &{0.881$\pm$0.003} &\textbf{0.903$\pm$0.002} &{0.891$\pm$0.003} \\
  ~ &CIFAR-10 &{0.734$\pm$0.007}$^{\ddagger}$ &{0.782$\pm$0.004}$^{\ddagger}$ &{0.791$\pm$0.004}$^{\ddagger}$ &{0.794$\pm$0.003}$^{\ddagger}$ &{0.767$\pm$0.004}$^{\ddagger}$ &{0.865$\pm$0.002}$^{\ddagger}$  &\textbf{0.887$\pm$0.002} \\
  \hline
  \multirow{2}{*}{Micro-F1} &FMNIST &{0.881$\pm$0.001}$^{\ddagger}$ &{0.891$\pm$0.002} &{0.892$\pm$0.002} &{0.885$\pm$0.004} &{0.882$\pm$0.004} &\textbf{0.903$\pm$0.002} &{0.892$\pm$0.002} \\
  ~ &CIFAR-10 &{0.735$\pm$0.008}$^{\ddagger}$ &{0.783$\pm$0.004}$^{\ddagger}$ &{0.791$\pm$0.004}$^{\ddagger}$ &{0.795$\pm$0.002}$^{\ddagger}$ &{0.768$\pm$0.004}$^{\ddagger}$ &{0.866$\pm$0.002}$^{\ddagger}$  &\textbf{0.887$\pm$0.002} \\
  \hline\hline
  \multicolumn{9}{c}{(\uppercase\expandafter{\romannumeral4}) FPS ($q=0.7$)} \\
  \hline
  \multirow{2}{*}{Macro-F1} &FMNIST &{0.874$\pm$0.007} &\textbf{0.886$\pm$0.006} &{0.884$\pm$0.003} &{0.875$\pm$0.002} &{0.863$\pm$0.002}$^{\ddagger}$ &{0.865$\pm$0.035}$^{\ddagger}$ &{0.878$\pm$0.005} \\
  ~ &CIFAR-10 &{0.678$\pm$0.008}$^{\ddagger}$ &{0.728$\pm$0.002}$^{\ddagger}$ &{0.745$\pm$0.004}$^{\ddagger}$ &{0.743$\pm$0.005}$^{\ddagger}$ &{0.673$\pm$0.051}$^{\ddagger}$ &{0.808$\pm$0.045}$^{\ddagger}$  &\textbf{0.869$\pm$0.006}  \\
  \hline
  \multirow{2}{*}{Micro-F1} &FMNIST &{0.875$\pm$0.007} &\textbf{0.886$\pm$0.002} &{0.885$\pm$0.001} &{0.875$\pm$0.001} &{0.853$\pm$0.002}$^{\ddagger}$ &{0.872$\pm$0.020}  &{0.879$\pm$0.004} \\
  ~ &CIFAR-10 &{0.681$\pm$0.008}$^{\ddagger}$ &{0.730$\pm$0.002}$^{\ddagger}$ &{0.746$\pm$0.004}$^{\ddagger}$ &{0.744$\pm$0.005}$^{\ddagger}$ &{0.692$\pm$0.035}$^{\ddagger}$ &{0.816$\pm$0.032}$^{\ddagger}$  &\textbf{0.870$\pm$0.006} \\
  \Xhline{1.5pt}
  \end{tabular}}
\end{table*}

\subsection{Experimental Setup}

\paragraph{Datasets.} 
We utilize 3 widely used benchmark image datasets, including Fashion-MNIST \cite{FashionMNIST2017}, CIFAR-10 and CIFAR-100 \cite{CIFAR2009}. We manually synthesize the partially labeled versions of these datasets by applying two data generation strategies, including Uniformly Sampling Strategy (USS) \cite{RCCC2020} and Flipping Probability Strategy (FPS) \cite{PRODEN2020}. The former one is conducted by uniformly sampling a candidate label set from the candidate label set space $\mathcal{C}$ for each instance, and the latter one generates the candidate label set of each instance by selecting any irrelevant label as its candidate one with a flipping probability $q$.\footnote{Note that the flipping probability strategy will uniformly flip a random irrelevant label into the candidate label set when none of irrelevant labels are flipped.} In experiments, we employ $q\in\{0.3,0.5,0.7\}$ for Fashion-MNIST and CIFAR-10, and $q\in\{0.05,0.1,0.2\}$ for CIFAR-100 due to the more labels. We adopt 5-layer LeNet, 22-layer Densenet and 18-layer ResNet as the backbones of Fashion-MNIST, CIFAR-10 and CIFAR-100, respectively. 

\begin{table*}[t]
  \centering
  \caption{Empirical results (mean$\pm$std) on CIFAR-100 with FPS ($q = 0.05, 0.1, 0.2$). The highest scores are indicated in \textbf{bold}. The notation ``$\ddagger$'' indicates that the performance gain of \baby is statistically significant (paired sample t-tests) at $0.01$ level.}
  \label{table2}
  \footnotesize
  \resizebox{0.98\linewidth}{!}{
  \begin{tabular}{p{36pt}<{\centering}|p{36pt}<{\centering}||p{48pt}<{\centering}p{48pt}<{\centering}p{48pt}<{\centering}p{48pt}<{\centering}p{48pt}<{\centering}p{48pt}<{\centering}p{48pt}<{\centering}}
  \Xhline{1.5pt}
  Metric &$q$ &\textbf{CC} &\textbf{RC} &\textbf{PRODEN} &\textbf{LWS} &\textbf{CAVL} &\textbf{PiCO}  &\textbf{\baby} \\
  \hline\hline
  \multirow{3}{*}{Macro-F1} &0.05  &{0.469$\pm$0.003}$^{\ddagger}$ &{0.461$\pm$0.007}$^{\ddagger}$ &{0.601$\pm$0.004}$^{\ddagger}$ &{0.567$\pm$0.008}$^{\ddagger}$ &{0.398$\pm$0.008}$^{\ddagger}$ &{0.744$\pm$0.007}$^{\ddagger}$ &\textbf{0.770$\pm$0.002} \\
  ~ &0.1  &{0.431$\pm$0.006}$^{\ddagger}$ &{0.388$\pm$0.006}$^{\ddagger}$ &{0.512$\pm$0.006}$^{\ddagger}$ &{0.498$\pm$0.005}$^{\ddagger}$ &{0.229$\pm$0.018}$^{\ddagger}$ &{0.636$\pm$0.021}$^{\ddagger}$ &\textbf{0.733$\pm$0.013} \\
  ~ &0.2  &{0.348$\pm$0.008}$^{\ddagger}$ &{0.230$\pm$0.013}$^{\ddagger}$ &{0.476$\pm$0.010}$^{\ddagger}$ &{0.401$\pm$0.015}$^{\ddagger}$ &{0.066$\pm$0.010}$^{\ddagger}$ &{0.190$\pm$0.025}$^{\ddagger}$ &\textbf{0.660$\pm$0.008} \\
  \hline
  \multirow{3}{*}{Micro-F1} &0.05  &{0.470$\pm$0.004}$^{\ddagger}$ &{0.465$\pm$0.006}$^{\ddagger}$ &{0.607$\pm$0.001}$^{\ddagger}$  &{0.596$\pm$0.003}$^{\ddagger}$ &{0.402$\pm$0.008}$^{\ddagger}$ &{0.746$\pm$0.006}$^{\ddagger}$  &\textbf{0.770$\pm$0.002} \\
  ~ &0.1  &{0.435$\pm$0.005}$^{\ddagger}$ &{0.400$\pm$0.005}$^{\ddagger}$ &{0.568$\pm$0.003}$^{\ddagger}$ &{0.535$\pm$0.001}$^{\ddagger}$ &{0.262$\pm$0.015}$^{\ddagger}$ &{0.660$\pm$0.015}$^{\ddagger}$ &\textbf{0.739$\pm$0.008} \\
  ~ &0.2  &{0.357$\pm$0.009}$^{\ddagger}$ &{0.279$\pm$0.012}$^{\ddagger}$ &{0.496$\pm$0.007}$^{\ddagger}$ &{0.434$\pm$0.011}$^{\ddagger}$ &{0.104$\pm$0.009}$^{\ddagger}$ &{0.288$\pm$0.022}$^{\ddagger}$ &\textbf{0.687$\pm$0.006} \\
  \Xhline{1.5pt}
  \end{tabular}}
\end{table*}

\paragraph{Baseline PL learning methods and training settings.}
We compare \baby against the following 6 existing deep PL learning methods, including RC \cite{RCCC2020}, CC \cite{RCCC2020}, PRODEN \cite{PRODEN2020}, LW \cite{LW2021} with sigmoid loss function, CAVL \cite{CAVL2022}, and PiCO \cite{PiCO2022}. We train all methods by using the SGD optimizer, and search the learning rate from $\{0.0001,0.001,0.01,0.05,0.1,0.5\}$ and the weight decay from $\{10^{-6},10^{-5},\cdots,10^{-1}\}$. For all baselines and the pre-training-stage of \baby, we set the batch size 256 for Fashion-MNIST and CIFAR-10, and 64 for CIFAR-100. For all baselines, we employ the default or suggested settings of hyper-parameters in their papers and released codes. For \baby, we use the following hyper-parameter settings: $\gamma_0=1.0,\;\lambda_0=0.01,\;\tau_0=0.75$, number of pre-training epoches $T_0=10$, number of SS training epoches $T=250$, number of inner loops $I=200$, batch sizes of pseudo-labeled and pseudo-unlabeled instances $B_l=64,B_u=256$. Specially, for CIFAR-100 we set $T_0=50,\;I=800,\;B_l=16,\;B_u=64$. We set the number of pseudo-labeled instances per-class $k=200$. Besides, we employ the horizontal flipping and cropping to conduct the weakly augmentation function $\alpha(\cdot)$ of all datasets, and implement the strongly augmentation function $\mathcal{A}(\cdot)$ for Fashion-MNIST with horizontal flipping, cropping and Cutout, for CIFAR-10 and CIFAR-100 with horizontal flipping, cropping, Cutout as well as AutoAugment.\footnote{For AutoAugment, we simply utilize the augmentation policies released by \cite{AutoAugment2019}.} 
All experiments are carried on a Linux server with one NVIDIA GeForce RTX 3090 GPU.

\paragraph{Evaluation metrics.}
We employ Macro-F1 and Micro-F1 to evaluate the classification performance, and calculate them by using the Scikit-Learn tools \cite{SKLEARN2011}.


\subsection{Main Results}
We perform all experiments with five different random seeds, and report the average scores of Fashion-MNIST and CIFAR-10 in Table \ref{table1}, and ones of CIFAR-100 in Table \ref{table2}. Overall, our \baby significantly outperforms all comparing methods in most cases, and achieves particularly significant performance gain on high ambiguity levels. As shown in Tables \ref{table1} and \ref{table2}: (1) Our \baby consistently perform better than all baselines on CIFAR-10 and achieves a competitive performance on Fashion-MNIST across four partial label settings. For example, Micro-F1 scores of \baby are $0.019\sim 0.054$ higher than ones of the recent state-of-the-art PiCO on four partially-labeled versions of CIFAR-10, and even gain $0.054$ significant improvement on high ambiguity level, \ie $q=0.7$. (2) Compared with all baselines, our \baby achieves very significant performance gain on CIFAR-100 across $q=0.05,0.1$ and $0.2$, and show more significant superiority than that on previous simpler datasets. (3) Besides, PiCO always drop dramatically on Fashion-MNIST and CIFAR-10 with $q=0.7$, especially CIFAR-100 with $q=0.2$. The possible reason is that PiCO could not identify true labels with contrastive representation learning to disambiguate candidate labels when on high ambiguity level.

\begin{table}[t]
  \centering
  \caption{Ablation study results (mean$\pm$std) on CIFAR-10 with FPS ($q=0.7$). The highest scores are indicated in \textbf{bold}.}
  \label{table3}
  \footnotesize
  \begin{tabular}{p{50pt}<{\centering}||p{50pt}<{\centering}p{50pt}<{\centering}}
  \Xhline{1.5pt}
  \multirow{2}{*}{Method} &\multicolumn{2}{c}{CIFAR-10}  \\
  \cline{2-3}
  ~ &Macro-F1 &Micro-F1  \\
  \hline
  \baby &\textbf{0.869$\pm$0.006} &\textbf{0.870$\pm$0.006}  \\
  \baby w/o ST &{0.858$\pm$0.008} &{0.861$\pm$0.007}  \\ 
  DF &{0.413$\pm$0.015} &{0.418$\pm$0.012}  \\ 
  \Xhline{1.5pt}
  \end{tabular}
\end{table}

\subsection{Ablation Study}
In this section, we perform extensive experiments to examine the importance of different components of \baby. We compare \baby with \baby without the semantic transformation (ST) and the version training only with the disambiguation-free (DF) objective of Eq.\eqref{eq-3-0} on CIFAR-10 by using data generation with FPS on $q=0.7$. The experimental results are reported in Table \ref{table3}. It clearly demonstrates that the proposed SS learning strategy can significantly improve the classification performance of PL learning. Besides, we can also observe that the semantic transformation can also improve the classification performance, proving its effectiveness to capture the semantic consistency.

\subsection{Sensitivity Analysis}
\begin{figure}[t]
  \includegraphics[width=0.45\textwidth]{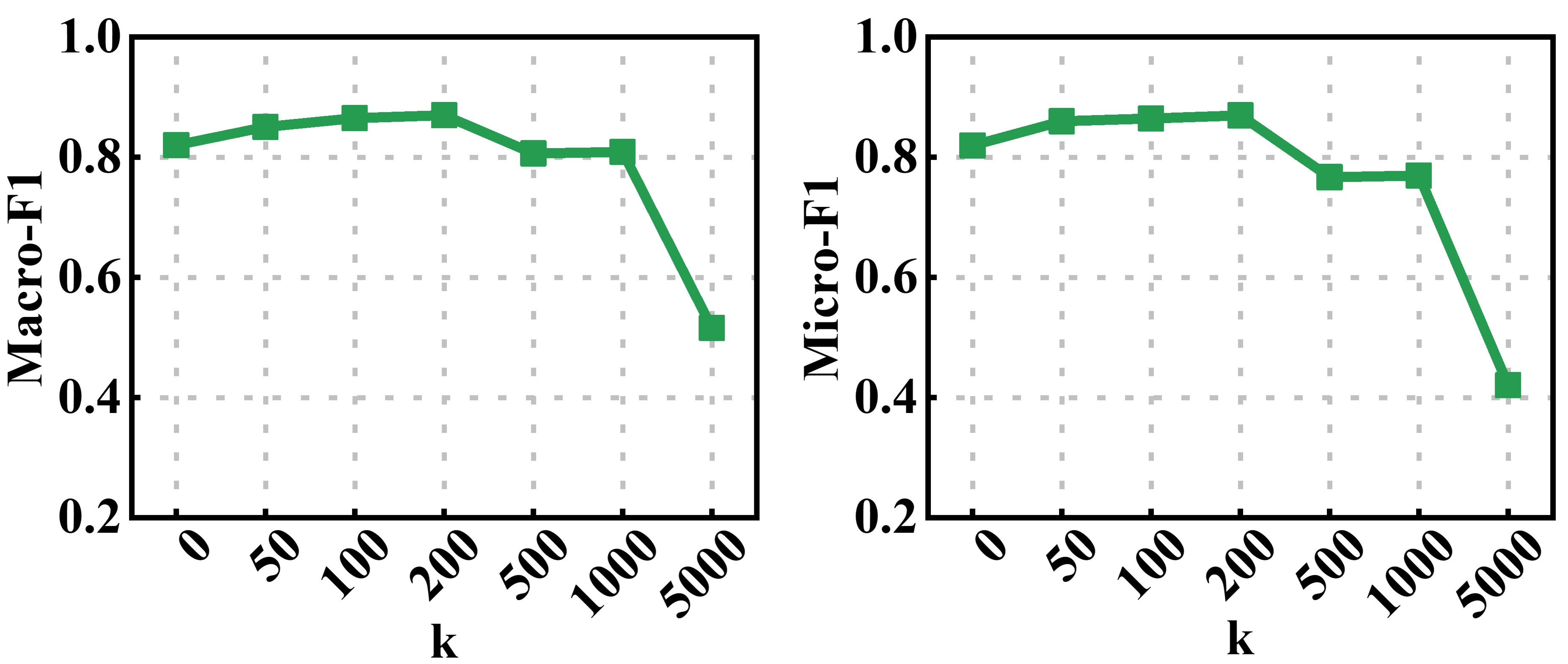}
  \centering
  \caption{Sensitivity analysis of the number of pseudo-labeled instances per-class $k$ on CIFAR-10 with FPS ($q=0.7$).}
  \label{fig-m}
\end{figure}

In this section, we examine the sensitivities of number of pseudo-instances per-class $k$. We conduct the sensitive experiments by varying $k$ over $\{0,50,100,200,500,1000,5000\}$ on CIFAR-10 by using data generation with FPS on $q=0.7$, and illustrate the experimental results in Fig.\ref{fig-m}. As is shown: (1) Obviously, the performance is relatively stable when $k\leq 200$ and achieve the highest when $k=200$, and it sharply drops as the values become bigger. It is expected since the smaller value of $k$ may ignore some high-confidence instances and the bigger value of $k$ will introduce many ``unreliable'' pseudo-labeled instances, leading to a poor classifier. (2) Moreover, the performance is poor when both $k=0$ and $5000$, especially when $k=5000$. Notice that when $k=0$ none of instances within $\Omega$ are selected as pseudo-labeled instances, \ie $\{\Omega_l=\emptyset,\Omega_u=\Omega\}$, and when $k=5000$ for CIFAR-10 all instances are selected as pseudo-labeled instances, \ie $\{\Omega_l=\Omega,\Omega_u=\Omega\}$. It demonstrates the effectiveness of the proposed SS learning strategy for PL learning task. In practice, we suggest tuning $k$ over the set $\{50, 100, 200\}$.

\begin{table}[t]
  \centering
  \caption{Time cost (second, \textbf{s}) of \baby and PiCO on Fashion-MNIST (FMNIST), CIFAR-10 and CIFAR-100 with USS.}
  \label{table4}
  \footnotesize
  \begin{tabular}{p{50pt}<{\centering}||p{30pt}<{\centering}p{30pt}<{\centering}|p{30pt}<{\centering}p{30pt}<{\centering}}
  \Xhline{1.5pt}
  \multirow{2}{*}{Method} &\multicolumn{2}{c|}{PiCO} &\multicolumn{2}{c}{\baby}\\
  \cline{2-5}
  ~ &Pretrain &Train &Pretrain &Train \\
  \hline
  FMNIST &-- &24,400s &60s &10,920s  \\
  CIFAR-10  &-- &38,000s &107s &25,000s  \\
  CIFAR-100 &-- &55,200s &299s &44,000s \\
  \Xhline{1.5pt}
  \end{tabular}
\end{table}

\subsection{Efficiency Comparison}
To examine the efficiency of our \baby, we perform efficiency comparisons over \baby and PiCO on all benchmarks with USS. We compare the overall time costs during pre-training and training stages respectively, and perform experiments with the suggested settings for all methods and benchmarks. Table \ref{table4} shows the running time results averaged on 10 runs. As is shown: (1) Obviously, the additional disambiguation-free pre-training stage of \baby is very efficient. (2) Moreover, In contrast to PiCO, \baby empirically converges fast due to more reliable supervision with SSL perspective (\baby 250 epochs vs PiCO 800 epochs) and costs less time in practice during the training stage.

\section{Conclusion}
\label{5}
In this work, we develop a novel PL learning method named \baby by resolving the PL learning problem with strong SS learning techniques. We conduct the SS learning strategy by selecting high-confidence partially-labeled instances as pseudo-labeled instances and treating the ones as pseudo-unlabeled. We design a semantic consistency regularization with respect to the semantic-transformed weakly- and strongly-augmented instances, and derive its approximation form for efficient optimization. Empirical results demonstrate the superior performance of \baby compared with the existing PL learning baselines, especially on high ambiguity levels.

\appendix

\section{Acknowledgments}
We would like to acknowledge support for this project from
the National Key R\&D Program of China (No.2021ZD0112501, No.2021ZD0112502),
the National Natural Science Foundation of China (NSFC) (No.62276113, No.62006094, No.61876071), the Key R\&D Projects of Science and Technology Department of Jilin Province of China (No.20180201003SF, No.20190701031GH).

\bibliography{PLLSSL}

\section{Proof of Proposition 1}

\begin{proof}
According to the definition of $\mathcal{R}_u^{\infty}((\mathbf{x}_i,C_i);\mathbf{\Theta})$ in Eq.(12), we have:
\begin{align}
    &\!\!\!\!\!\mathcal{R}_u^{\infty}((\mathbf{x}_i,C_i);\mathbf{\Theta}) \nonumber \\
    &\!\!\!\!\!=\mathbb{E}_{\underline{\mathbf{\widehat a}}_i^{w,k_1},\underline{\mathbf{ a}}_i^{s,k_2}}[h(\mathbf{\widehat p}_i^{w,k_1})\text{KL}(\mathbf{\widehat p}_i^{k_1}||\mathbf{p}_i^{s,k_2})] \nonumber \\
    &\!\!\!\!\!=h(\mathbb{E}_{\underline{\mathbf{\widehat a}}_i^{w,k_1}}[\mathbf{\widehat p}_i^{w,k_1}])\sum_{j\in\mathcal{Y}}\mathbb{E}_{\underline{\mathbf{\widehat a}}_i^{w,k_1},\underline{\mathbf{ a}}_i^{s,k_2}}[-\widehat{p}_{ij}^{k_1}\log p_{ij}^{s,k_2}] \nonumber \\
    &\!\!\!\!\!=h(\mathbb{E}_{\underline{\mathbf{\widehat a}}_i^{w,k_1}}[\mathbf{\widehat p}_i^{w,k_1}])\sum_{j\in\mathcal{Y}}\mathbb{E}_{\underline{\mathbf{\widehat a}}_i^{w,k_1}}[\widehat{p}_{ij}^{k_1}]\mathbb{E}_{\underline{\mathbf{ a}}_i^{s,k_2}}[-\log p_{ij}^{s,k_2}], \label{eq-a-0}
\end{align}
where $\mathbb{E}_{\underline{\mathbf{\widehat a}}_i^{w,k_1}}[\mathbf{\widehat p}_i^{w,k_1}]=[\mathbb{E}_{\underline{\mathbf{\widehat a}}_i^{w,k_1}}[\widehat{p}_{ij}^{w,k_1}]]_{j\in\mathcal{Y}}^{\top}$,
\begin{equation*}
    \mathbb{E}_{\underline{\mathbf{\widehat a}}_i^{w,k_1}}[\widehat{p}_{ij}^{k_1}]=\frac{\mathds{1}(j\in C_i)\mathbb{E}_{\underline{\mathbf{\widehat a}}_i^{w,k_1}}[\widehat{p}_{ij}^{w,k_1}]}{\sum_{j'\in \mathcal{Y}}\mathds{1}(j\in C_i)\mathbb{E}_{\underline{\mathbf{\widehat a}}_i^{w,k_1}}[\widehat{p}_{ij'}^{w,k_1}]}.
\end{equation*}
Here, we neglect $\sum_{j\in\mathcal{Y}}\mathbb{E}_{\underline{\mathbf{\widehat a}}_i^{w,k_1}}[\widehat{p}_{ij}^{k_1}\log\widehat{p}_{ij}^{k_1}]$ in the KL-divergence since it is a constant with respect to the classifier parameters $\mathbf{\Theta}$. In the following, we upper bound the expectations $\mathbb{E}_{\underline{\mathbf{\widehat a}}_i^{w,k_1}}[\widehat{p}_{ij}^{w,k_1}]$ and $\mathbb{E}_{\underline{\mathbf{ a}}_i^{s,k_2}}[-\log p_{ij}^{s,k_2}]$, respectively.

We approximate the expectation $\mathbb{E}_{\underline{\mathbf{\widehat a}}_i^{w,k_1}}[\widehat{p}_{ij}^{w,k_1}]$ as below:
\begin{align}
&\mathbb{E}_{\underline{\mathbf{\widehat a}}_i^{w,k_1}}[\widehat{p}_{ij}^{w,k_1}] =\mathbb{E}_{\underline{\mathbf{\widehat a}}_i^{w,k_1}}\biggl[\frac{e^{\mathbf{\widehat w}_j^{\top}\underline{\mathbf{\widehat a}}_i^{w,k_1}}}{\sum_{j'\in\mathcal{Y}}e^{\mathbf{\widehat w}_{j'}^{\top}\underline{\mathbf{\widehat a}}_i^{w,k_1}}}\biggr] \nonumber \\
&=\mathbb{E}_{\underline{\mathbf{\widehat a}}_i^{w,k_1}}\biggl[\frac{1}{\sum_{j'\in\mathcal{Y}}e^{-\mathbf{\widehat u}_{jj'}^{\top}\underline{\mathbf{\widehat a}}_i^{w,k_1}}}\biggr] \nonumber \\
&=\mathbb{E}_{\underline{\mathbf{\widehat a}}_i^{w,k_1}}\biggl[\frac{1}{-l+\sum_{j'\in\mathcal{Y}}1+e^{-\mathbf{\widehat u}_{jj'}^{\top}\underline{\mathbf{\widehat a}}_i^{w,k_1}}}\biggr] \nonumber \\
&=\mathbb{E}_{\underline{\mathbf{\widehat a}}_i^{w,k_1}}\biggl[\frac{1}{-l+\sum_{j'\in\mathcal{Y}}1/\mathfrak{s}(\mathbf{\widehat u}_{jj'}^{\top}\underline{\mathbf{\widehat a}}_i^{w,k_1})}\biggr] \nonumber \\
&\approx \frac{1}{-l+\sum_{j'\in\mathcal{Y}}1/\mathbb{E}_{\underline{\mathbf{\widehat a}}_i^{w,k_1}}[\mathfrak{s}(\mathbf{\widehat u}_{jj'}^{\top}\underline{\mathbf{\widehat a}}_i^{w,k_1})]} \nonumber \\
&\approx \frac{1}{-l+\sum_{j'\in\mathcal{Y}}1/\Phi\Bigl(\frac{\beta\mathbf{\widehat u}_{jj'}^{\top}\mathbf{\widehat a}_i^w}{(1+\lambda\beta^2\mathbf{\widehat u}_{jj'}^{\top}\mathbf{\Sigma}_{\widehat{y}_i}\mathbf{\widehat u}_{jj'})^{1/2}}\Bigr)}=\underline{\widehat{p}}_{ij}^w,
\label{eq-a-1}
\end{align}
where $\mathbf{\widehat u}_{jj'}=\mathbf{\widehat w}_j-\mathbf{\widehat w}_{j'}$, $\mathfrak{s}(a)\coloneqq(1+e^{-a})^{-1}$ is the sigmoid function, and $\Phi(z)=\frac{1}{\sqrt{2\pi}}\int_{-\infty}^ze^{-t^2/2}dt$ is the cumulative distribution function of the standard normal distribution $\mathcal{N}(0,1)$. In the above, Eq.\eqref{eq-a-1} is obtained by leveraging:
\begin{align*}
\mathbb{E}[\mathfrak{s}(X)] &= \int_X \mathfrak{s}(X)\mathcal{N}(X;\mu,\sigma^2) dX \\
&\approx \int_X\Phi(\beta X)\mathcal{N}(X;\mu,\sigma^2)dX=\Phi(\frac{\beta\mu}{\sqrt{1+\beta^2\sigma^2}})
\end{align*}
for some fine-tuned $\beta>0$ (\eg $\beta=\pi^2/8$) due to the fact that $\mathbf{\widehat u}_{jj'}^{\top}\underline{\mathbf{\widehat a}}_i^{w,k_1}$ is a Gaussian random variable:
\begin{equation*}
    \mathbf{\widehat u}_{jj'}^{\top}\underline{\mathbf{\widehat a}}_i^{w,k_1}\sim\mathcal{N}(\mathbf{\widehat u}_{jj'}^{\top}\mathbf{\widehat a}_i^{w,k_1},\lambda\mathbf{\widehat u}_{jj'}^{\top}\mathbf{\Sigma}_{\widehat{y}_i}\mathbf{\widehat u}_{jj'}).
\end{equation*}

Next, according to the calculation of $p_{ij}^{s,k_2}$, we upper bound the expectation $\mathbb{E}_{\underline{\mathbf{ a}}_i^{s,k_2}}[-\log p_{ij}^{s,k_2}]$ as follows: 
\begin{align}
&\mathbb{E}_{\underline{\mathbf{ a}}_i^{s,k_2}}[-\log p_{ij}^{s,k_2}]=\mathbb{E}_{\underline{\mathbf{ a}}_i^{s,k_2}}\biggl[-\log \frac{e^{\mathbf{w}_j^{\top}\underline{\mathbf{ a}}_i^{s,k_2}}}{\sum_{j'\in\mathcal{Y}}e^{\mathbf{w}_{j'}^{\top}\underline{\mathbf{ a}}_i^{s,k_2}}}\biggr] \nonumber \\
&=\mathbb{E}_{\underline{\mathbf{ a}}_i^{s,k_2}}\biggl[\log(\sum_{j'\in\mathcal{Y}}e^{\mathbf{u}_{j'j}^{\top}\underline{\mathbf{ a}}_i^{s,k_2}})\biggr] \nonumber \\
&\leq \log\biggl(\sum_{j'\in\mathcal{Y}}\mathbb{E}_{\underline{\mathbf{ a}}_i^{s,k_2}}\Bigl[e^{\mathbf{u}_{j'j}^{\top}\underline{\mathbf{ a}}_i^{s,k_2}}\Bigr]\biggr) \label{eq-a-2-1} \\
&=\log\biggl(\sum_{j'\in\mathcal{Y}}e^{\mathbf{u}_{j'j}^{\top}\mathbf{a}_i^s+\frac{\lambda}{2}\mathbf{u}_{j'j}^{\top}\mathbf{\Sigma}_{\widehat{y}_i}\mathbf{u}_{j'j}}\biggr) \label{eq-a-2-2} \\
&=-\log\biggl(\frac{e^{\mathbf{w}_j^{\top}\mathbf{a}_i^s}}{\sum_{j'\in\mathcal{Y}}e^{\mathbf{w}_{j'}^{\top}\mathbf{a}_i^s+\frac{\lambda}{2}\mathbf{u}_{j'j}^{\top}\mathbf{\Sigma}_{\widehat{y}_i}\mathbf{u}_{j'j}}}\biggr)=-\log \underline{p}_{ij}^s, \label{eq-a-2}
\end{align}
where $\mathbf{u}_{j'j}=\mathbf{w}_{j'}-\mathbf{w}_j$. In the above, the inequality \eqref{eq-a-2-1} follows from the Jensen's inequality $\mathbb{E}[\log X]\leq\log(\mathbb{E}[X])$; Eq.\eqref{eq-a-2-2} is obtained by using the moment-generating function of the Gaussian distribution:
\begin{equation*}
\mathbb{E}[e^{tX}]=e^{t\mu+\frac{1}{2}\sigma^2t^2},\quad X\sim\mathcal{N}(\mu,\sigma^2),
\end{equation*}
due to the fact that $\mathbf{u}_{j'j}^{\top}\underline{\mathbf{ a}}_i^{s,k_2}$ is a Gaussian random variable:
\begin{equation*}
\mathbf{u}_{j'j}^{\top}\underline{\mathbf{ a}}_i^{s,k_2}\sim\mathcal{N}(\mathbf{u}_{j'j}^{\top}\mathbf{a}_i^s,\lambda\mathbf{u}_{j'j}^{\top}\mathbf{\Sigma}_{\widehat{y}_i}\mathbf{u}_{j'j}).
\end{equation*}

Accordingly, Proposition 1 is derived by combining Eqs.\eqref{eq-a-0}, \eqref{eq-a-1} and \eqref{eq-a-2}.

\end{proof}

\end{document}